\documentclass[10pt,twocolumn,letterpaper]{article}

\usepackage[pagenumbers]{cvpr} 

\definecolor{cvprblue}{rgb}{0.21,0.49,0.74}
\usepackage[pagebackref,breaklinks,colorlinks,allcolors=cvprblue]{hyperref}
\usepackage{microtype}
\usepackage{graphicx}
\usepackage{caption}
\usepackage{subcaption}
\usepackage{booktabs} 
\usepackage{tabularx}
\usepackage{multirow}
\usepackage{threeparttable}
\usepackage{siunitx}
\usepackage{amsmath}
\usepackage{mathtools}
\usepackage[export]{adjustbox}
\usepackage{tikz}
\usepackage{balance}
\usepackage{standalone}
\usetikzlibrary{positioning,arrows.meta}
\usepackage{siunitx}
\usepackage[dvipsnames]{xcolor}
\newcolumntype{Y}{>{\centering\arraybackslash}X}
\def\paperID{44366} 
\def\confName{CVPR}
\def\confYear{2026}
\newcommand{\Blueprint}{\textnormal{\textsc{Blueprint}}}

\title{\textsc{Blueprint} -- Rebuilding a Legacy: Multimodal Retrieval for Complex Engineering Drawings and Documents}

\author{
Ethan Seefried$^{1,2}$ \quad
Ran Eldegaway$^{1}$ \quad
Sanjay Das$^{1}$ \quad
Nathaniel Blanchard$^{2}$ \quad
Tirthankar Ghosal$^{1}$\\[0.5em]
$^{1}$Oak Ridge National Laboratory, Oak Ridge, Tennessee\\
$^{2}$Colorado State University, Fort Collins, Colorado\\[0.5em]
{\tt\small seefriedej@ornl.gov, ghosalt@ornl.gov}
}

\begin{document}

\maketitle

\begin{abstract}
Decades of engineering drawings and technical records remain locked in legacy archives with inconsistent or missing metadata, making retrieval difficult and often manual. We present \Blueprint{}, a layout-aware multimodal retrieval system designed for large-scale engineering repositories. \Blueprint{} detects canonical drawing regions, applies region-restricted VLM-based OCR, normalizes identifiers (e.g., DWG, part, facility), and fuses lexical and dense retrieval with a lightweight region-level reranker. Deployed on \(\sim\)770k unlabeled files, it automatically produces structured metadata suitable for cross-facility search.

We evaluate \Blueprint{} on a 5k-file benchmark with 350 expert-curated queries using pooled, graded (0/1/2) relevance judgments. \textbf{\textup{\textsc{Blueprint}} delivers a 10.1\% absolute gain in Success@3 and an 18.9\% relative improvement in nDCG\@3 over the strongest vision-language baseline}, consistently outperforming across vision, text, and multimodal intents. Oracle ablations reveal substantial headroom under perfect region detection and OCR. We release all queries, runs, annotations, and code to facilitate reproducible evaluation on legacy engineering archives.
\end{abstract}

\section{Introduction}

Engineering organizations rely on diagrams, drawings, and linked documentation to design, operate, and maintain complex systems \cite{murthy1985survey, ondrejcek2009information}. Over decades, these assets accumulate across heterogeneous and siloed repositories (e.g., shared drives, PLM/EDRM systems, vendor portals), with idiosyncratic authoring styles and inconsistent conventions \cite{waller2013makes}. As sites modernize digital infrastructure and senior experts retire, critical knowledge risks becoming inaccessible \cite{burmeister2016knowledge, daghfous2013understanding, massingham2018measuring, bell2014mitigating, huet2007knowledge, robillard2021turnover}. Fully re-cataloging legacy archives would require annotating hundreds of thousands of files, costly, slow, and error-prone, making scalable \emph{retrieval} rather than re-labeling the core challenge.

\begin{figure}[t]
\centering
\resizebox{\linewidth}{!}{\input{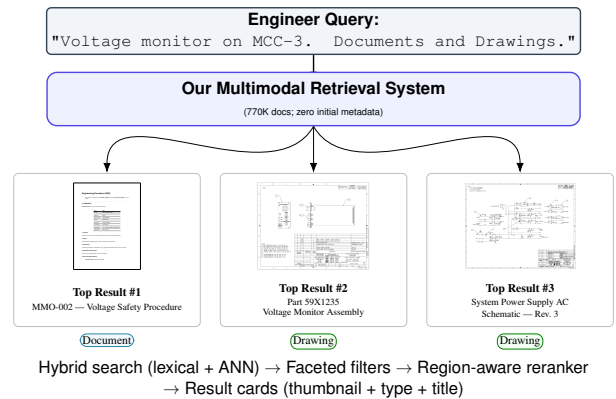}}
\caption{\textsc{Blueprint} processes an engineer’s natural-language query and returns multimodal results across drawings, schematics, and procedural documents.}
\end{figure}

Legacy corpora are heterogeneous in format (vector CAD, raster scans, TIFF/PDF, mixed-mode), quality (low-resolution scans, skew, bleed-through), and metadata (often missing or unreliable). Collections are inherently \emph{multi-modal}: visual drawings co-exist with policies, procedures, and narrative reports. Engineers query in text, but evidence often resides in visual regions such as title blocks, revision tables, or symbols. OCR noise, domain abbreviations, and near-duplicate revisions further complicate retrieval \cite{bazzo2020assessing,most2025lost}. These constraints motivate systems that fuse sparse textual signals with layout- and region-aware visual cues for robust, scalable retrieval over engineering documents.

We propose \textsc{Blueprint}: Multimodal Retrieval for Engineering Archives, a framework that unifies computer vision and NLP for legacy engineering corpora. A lightweight \emph{modality router} sends each file down a vision-first path (drawings, scans) or a text-first path (policies, procedures). On the vision path, the system detects layout regions (e.g., title block, revisions, parts list), applies region-restricted OCR, and parses identifiers into structured metadata (e.g., drawing number, revision, facility tags, parts-list entries). On the text path, \textsc{Blueprint} extracts key entities and section-level structure from procedural documents. These signals are fused into a joint embedding and indexed in a \emph{hybrid} sparse+dense store. At query time, the system retrieves candidates via hybrid search and applies a lightweight reranker that promotes region-supported hits. The result is an efficient, end-to-end multimodal retrieval pipeline resilient to OCR noise, sparse text, and heterogeneous formats.

On a 5k-file benchmark with 375 mixed-modality queries, \textsc{Blueprint} delivers state-of-the-art retrieval performance, with $\text{Success}@3 = 0.715 \pm 0.150$ and $nDCG@3 = 0.626 \pm 0.146$, achieving $\approx 10\%$ absolute gains over the strongest VLM baseline. At $\approx 9.7\,\mathrm{s/file}$ end-to-end throughput, \textsc{Blueprint} is also the fastest. Pairwise LLM-as-judge comparisons show a $72.88\%$ win rate against leading VLMs.

Our main contributions are:
\begin{itemize}
    \item We introduce \textsc{Blueprint}, an end-to-end system that integrates region detection, region-restricted VLM OCR, identifier normalization, and hybrid sparse+dense retrieval with a lightweight region-level reranker. Deployed at scale over hundreds of thousands of unlabeled engineering files, \textsc{Blueprint} enables accurate cross-modal search across drawings, schematics, and technical documents.

      \item On a curated benchmark of 375 queries, \textsc{Blueprint} consistently outperforms all vision-language baselines, delivering clear improvements in retrieval quality. This demonstrates that task-specific pipelines (e.g., YOLO + VLM OCR, LLM document routing) outperform general-purpose large VLMs such as LLaVA, Pixtral, and PaLI-Gemma.
    
    \item Region-level extraction substantially boosts retrieval accuracy, hybrid sparse+dense search outperforms single-modality retrieval, and OCR/box ablations reveal further headroom from improved layout understanding.
    
    \item We also contribute a graded relevance benchmark evaluated with three independent LLM judges and validated by human experts. \textsc{Blueprint} achieves a strong win rate against all baselines and establishes a scalable evaluation protocol for large engineering-document collections.
    
\end{itemize}

\section{Related Work}

\textbf{Document Retrieval Systems.}
Early retrieval systems relied on probabilistic and vector-space models \cite{croft1983document,croft1987i3r}, with tf–idf and BM25 establishing strong lexical baselines \cite{gery2012bm25t,robertson2009probabilistic}. Classic improvements, including Rocchio-style relevance feedback \cite{rocchio71relevance} and inverted-index optimizations \cite{baranchuk2018revisitinginvertedindicesbillionscale}, enabled efficient keyword search at scale. Learning-to-rank approaches (RankSVM, LambdaMART) \cite{cao2006adapting,burges2010ranknet} introduced supervised ranking over hand-crafted features.

Dense retrieval emerged with neural encoders such as DPR \cite{karpukhin2020densepassageretrievalopendomain}, which map queries and passages into a shared embedding space for MIPS search with FAISS \cite{johnson2019}. Subsequent work improved negative mining, cross-encoder distillation \cite{qu2021rocketqa,hofstätter2021efficiently}, and late-interaction architectures such as ColBERT \cite{khattab2020colbert,santhanam2021colbertv2}. LLM-based retrieval extends these ideas to text-generation settings, including RAG \cite{arslan2024survey,lewis2020retrieval}, reranking \cite{sun2024chatgptgoodsearchinvestigating}, and zero-shot retrieval \cite{shen-etal-2024-retrieval,yang2024ldre}. However, these systems remain text-centric and degrade in symbol-heavy, OCR-noisy domains like engineering drawings.

\textbf{Multimodal Retrieval for Technical Archives.}
Vision–language pretraining enables cross-modal matching through joint contrastive embeddings, as in CLIP/ALIGN \cite{radford2021learningtransferablevisualmodels, jia2021scalingvisualvisionlanguagerepresentation}, which perform strongly on natural-image retrieval benchmarks such as COCO, Flickr30K, and VirTex \cite{lin2015microsoftcococommonobjects, plummer2016flickr30kentitiescollectingregiontophrase, desai2021virtexlearningvisualrepresentations}. Document-understanding models—LayoutLM, DocFormer, LayoutLLM \cite{xu2020layoutlm, xu2020layoutlmv2, huang2022layoutlmv3, appalaraju2021docformer, luo2024layoutllm}—fuse OCR text, layout structure, and appearance, showing strong results on FUNSD, DocLayNet, PubTables-1M, InfographicVQA, and ICDAR tasks \cite{jaume2019funsddatasetformunderstanding, doclaynet2022, smock2021pubtables1mcomprehensivetableextraction, mathew2022infographicvqa, harley2015icdar}. These datasets assume relatively regular layouts and reliable OCR, which break down in legacy engineering archives.

Prior work in technical-domain retrieval targets scientific figure search \cite{siegel2016figureseer}, educational diagrams, or CAD retrieval, but engineering repositories pose additional challenges due to symbol-heavy layouts, facility-specific conventions, and the need for text–visual co-retrieval (e.g., retrieving both a safety policy and its referenced schematic). General VLMs (GPT-4V, LLaVA, Pixtral, Gemma-2) \cite{liu2023visualinstructiontuning, chu2024visionllama, zhang2023gpt, agrawal2024pixtral, team2025gemma, grattafiori2024llama} process pages holistically but struggle to recover schema-specific fields such as drawing numbers or revision identifiers. Traditional pipelines rely on OCR or detect–then-extract models \cite{kashnevik, villena2023optical, yao2022intelligent, sohan2024review, nakshathri}, while OCR-free approaches (Donut, Florence-2) \cite{khan2025multi, khan2025drawings, khan2024fine, kim2021donut, xiao2024florence} still underperform on irregular schematics. \textsc{Blueprint} closes this gap via region detection, VLM-OCR, structured parsing, and unified BM25+ANN indexing for schema-precise cross-modal retrieval at archive scale.

\textbf{Evaluation of Retrieval Systems.}
IR evaluation typically uses pooled judging \cite{voorhees2005trec}, graded relevance \cite{ir_retrieval, jarvelin20amp}, and multimodal benchmarks such as Flickr30K and COCO \cite{plummer2016flickr30kentitiescollectingregiontophrase, lin2015microsoftcococommonobjects}. Engineering queries lack ground-truth pairs, motivating pooled graded assessment. LLM-as-judge protocols \cite{zheng2023judgingllmasajudgemtbenchchatbot, dubois2025lengthcontrolledalpacaevalsimpleway} offer scalability but require validation. We adopt a pooled 0/1/2 scheme judged by three independent LLMs (GPT-5 \cite{openai2025gpt5}, LLaMA-3.2 \cite{grattafiori2024llama}, Mistral-7B \cite{jiang2023mistral7b}) with human auditing via Cohen’s~\(\kappa\) and Kendall’s~\(\tau\).

\begin{figure*}[t]
    \centering
    \begin{subfigure}{0.32\textwidth}
        \centering
        \includegraphics[frame,height=4.2cm,keepaspectratio,valign=c]{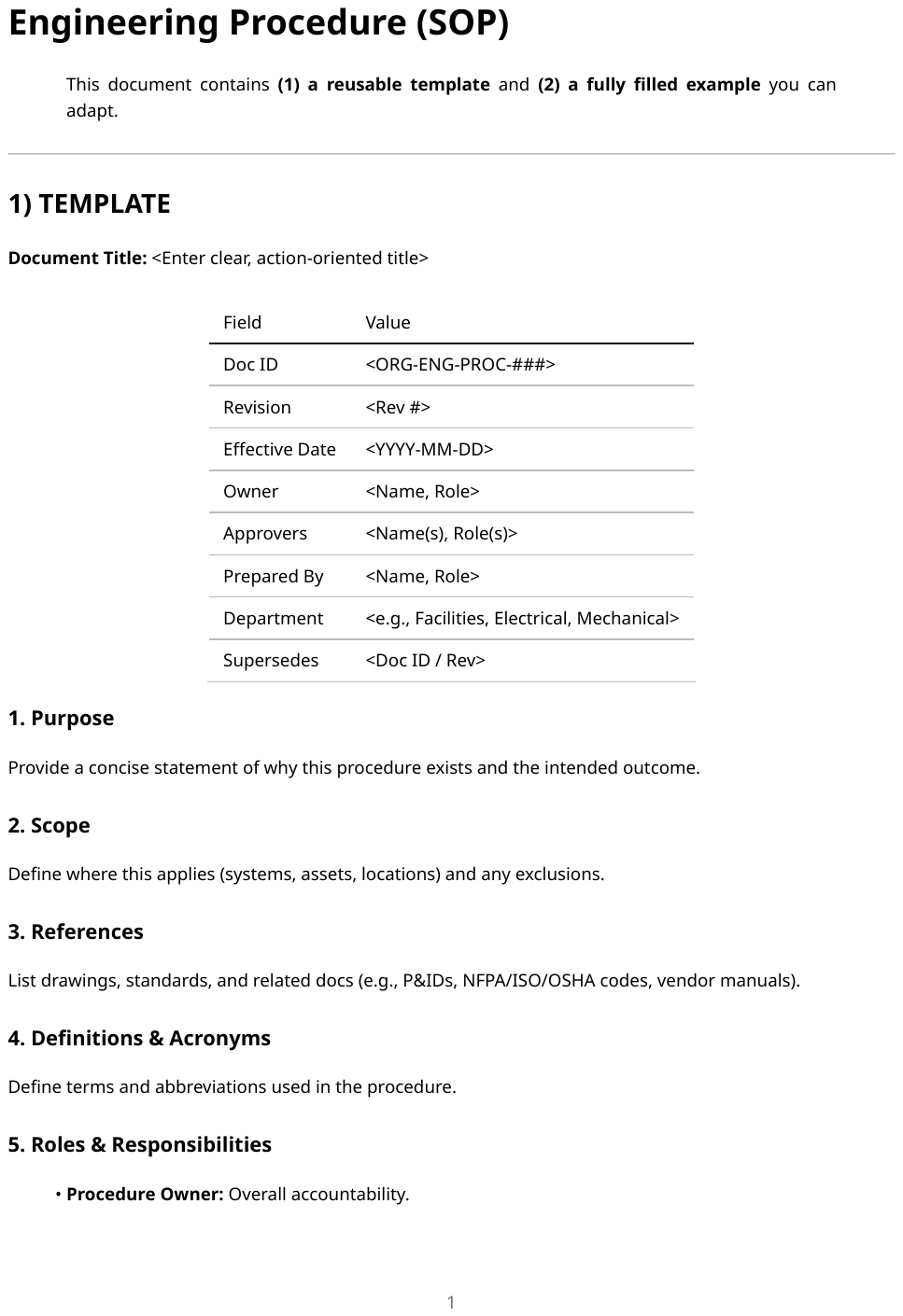}
        \caption{Procedure document.}
        \label{fig:example1}
    \end{subfigure}
    \hfill
    \begin{subfigure}{0.32\textwidth}
        \centering
        \includegraphics[width=.75\linewidth,rotate=90]{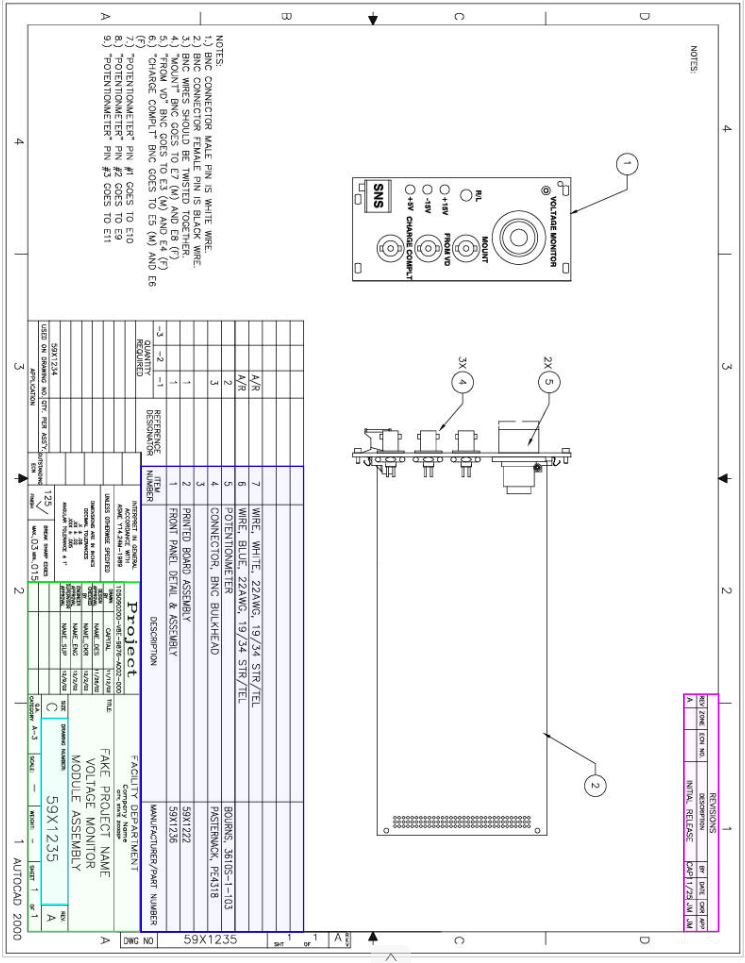}
        \caption{Assembly drawing.}
        \label{fig:example2}
    \end{subfigure}
    \hfill
    \begin{subfigure}{0.32\textwidth}
        \centering
        \includegraphics[width=.75\linewidth,rotate=90]{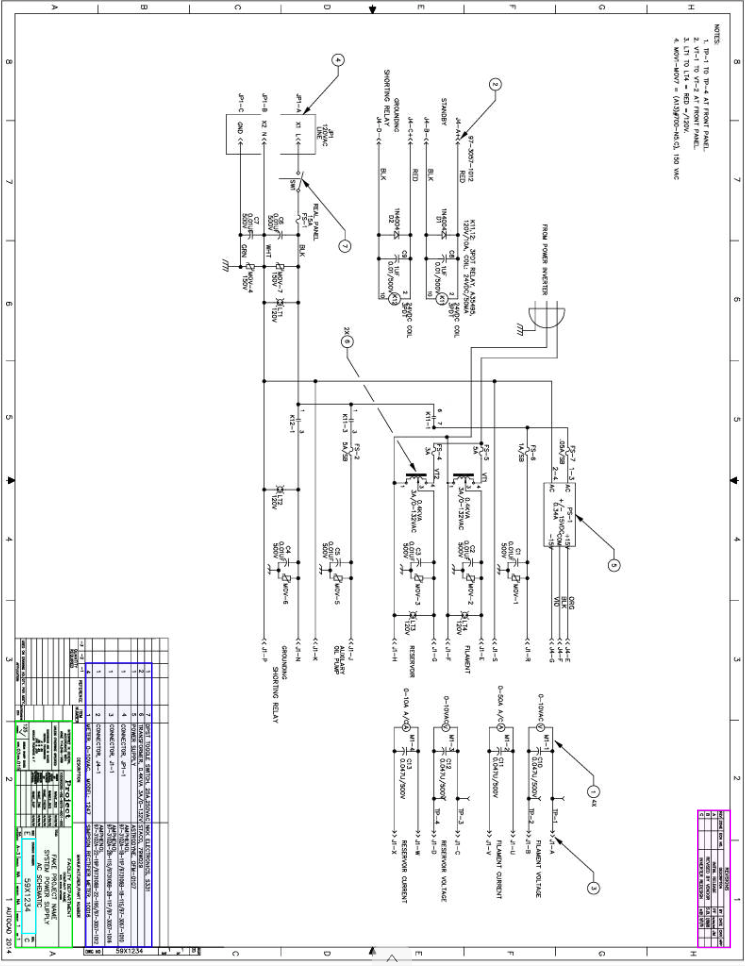}
        \caption{Electrical schematic.}
        \label{fig:example3}
    \end{subfigure}
    \caption{Representative samples: (a) policy/procedure with structured text, (b) assembly drawing with parts list and annotations, (c) electrical schematic with connection diagrams. Despite facility-specific conventions and missing regions in some drawings, common metadata areas (title block, drawing number, parts list, revisions) enable structured, layout-aware extraction for retrieval.}
    \label{fig:dataset_examples}
\end{figure*}

\begin{table}[th!]
\centering
\small
\caption{File type distribution in the original dataset before and after filtering.}
\label{tab:file-type-dist}
\setlength{\tabcolsep}{6pt}
\begin{tabular}{lrrr}
\toprule
File Type & Count & Percent (\%) & Extensions\\
\midrule
Document  & 327{,}303 & 45.01 & .pdf, .docx\\
Image     & 196{,}260 & 26.99 & .jpg, .tiff\\
Other     & 195{,}206 & 26.85 & .cad, .sldprt\\
Archive   & 7{,}392   & 1.02  & .zip, .tar\\
Video     & 930       & 0.13  & .mp4, .mov\\
Audio     & 28        & 0.00  & .mp3, .wav\\
\midrule
Engineering Drawings & 166{,}125 & 29.72 & .pdf, .sldprt\\
Documents            & 392{,}801 & 70.28 & .pdf, .docx\\
\bottomrule
\end{tabular}
\end{table}

\section{Dataset}\label{sec/dataset}
Our dataset consists of 770{,}000 unique files from a legacy document 
management system. Files contained limited or no metadata and spanned 
diverse formats: images, engineering drawings, textual documents, videos, 
and compressed archives. Since our goal is cross-modal retrieval between 
technical drawings and procedural documentation, we filtered for policies, 
procedures, and engineering drawings, yielding 558{,}926 unlabeled files.

\subsection{File Types}
Left with 558{,}926 unlabeled files, we primarily cared about 2 overarching categories of files: documents, and engineering drawings. Table \ref{tab:file-type-dist} displays the initial dataset before metadata extraction.

\textbf{Documents.} The primary textual materials of interest were organizational \emph{policies} and \emph{procedures}. However, the legacy database also included a range of unrelated files such as images, videos, audio recordings, and archives. Document formatting was inconsistent across facilities, though most were stored as PDF or Microsoft Word files. In total, 392{,}801 documents were retained. Figure~\ref{fig:example1} illustrates an example procedure document, where the document should be templated.

\textbf{Engineering Drawings.}
Engineering drawings in the dataset originated from multiple \emph{anonymous 
facilities}, each employing distinct design conventions and labeling schemes. 
\textbf{Drawings from Facility A use metric units and DIN standards, while 
Facility B follows ANSI conventions, yet our retrieval system must handle 
both without facility-specific customization.} As a result, no unified 
structure existed for extracting information. The drawings spanned a wide spectrum, from small parts to complete assemblies and electrical schematics, and appeared under various file extensions (e.g., PDF, CAD, etc.). In total, 392{,}801 documents were retained. Figures~\ref{fig:example2}, and \ref{fig:example3}
show example engineering diagrams and schematics that are related to each other.

\subsection{Annotations}
\textbf{Test Sets.} To support model development and evaluation, we constructed two manually annotated subsets from the 770{,}000 unlabeled files. First, a 500-sample pilot set (211 drawings, 289 documents) was randomly selected and labeled using a lightweight \texttt{matplotlib}-based interface; this enabled rapid iteration and early model selection. Building on this, we assembled a higher-confidence \emph{golden test set} of 1{,}500 new, non-overlapping files, sampled at a balanced 50/50 split (750 drawings, 750 documents) and annotated using the same procedure. The golden set serves as the primary benchmark for comparing tuned models.

\textbf{Metadata Extraction.}\label{metadata}
We annotated the 750-image golden test set using CVAT~\cite{cvat2023}, producing YOLO-format bounding boxes for four standardized metadata regions commonly found in engineering drawings: \emph{drawing number}, \emph{data block}, \emph{parts list}, and \emph{revisions block}. Of the 750 drawings, 749 were successfully labeled. These regions were selected because they contain structured identifiers critical for indexing and retrieval.

Table~\ref{tab:dataset-splits} reports the train/val/test splits along with counts of each metadata region. Coverage is high across all splits, with only a handful of drawings missing individual fields. Splits were assigned randomly with the constraint that each subset contained a representative number of \emph{parts lists}, ensuring balanced supervision for this key region. This labeled set provides a reliable foundation for training and evaluating region detectors whose accuracy directly influences downstream retrieval quality.


\begin{table}[t]
\centering
\caption{Summary of dataset splits showing the number of drawings and labeled metadata regions (parts list, data block, drawing number, and revisions block) across train, validation, and test sets.}
\begin{tabular*}{\columnwidth}{@{\extracolsep{\fill}}lrrrrr}
\toprule
Split & Files & Parts & Data & Drw. & Rev. \\
\midrule
Train & 500 & 232 & 495 & 496 & 455 \\
Val   & 150 & 70  & 150 & 150 & 150 \\
Test  & 99  & 45  & 98  & 98  & 89  \\
\bottomrule
\end{tabular*}

\label{tab:dataset-splits}
\end{table}

\textbf{Dataset availability.} The operational corpus contains sensitive engineering records and cannot be released. We provide full code, prompts, normalization rules, index configurations, seeds. To enable reproducibility, we will release an independent 5k+ benchmark of unrestricted engineering documents, with the same queries and relevance judgments, which will serve as the held-out test set for an upcoming shared task. A larger 50k+ open dataset of annotated engineering diagrams (components, relationships, multimodal tasks) is also in development and will be released with baselines, culminating in a community shared task.

\begin{figure*}[h!t]
    \centering
    \includegraphics[width=\textwidth]{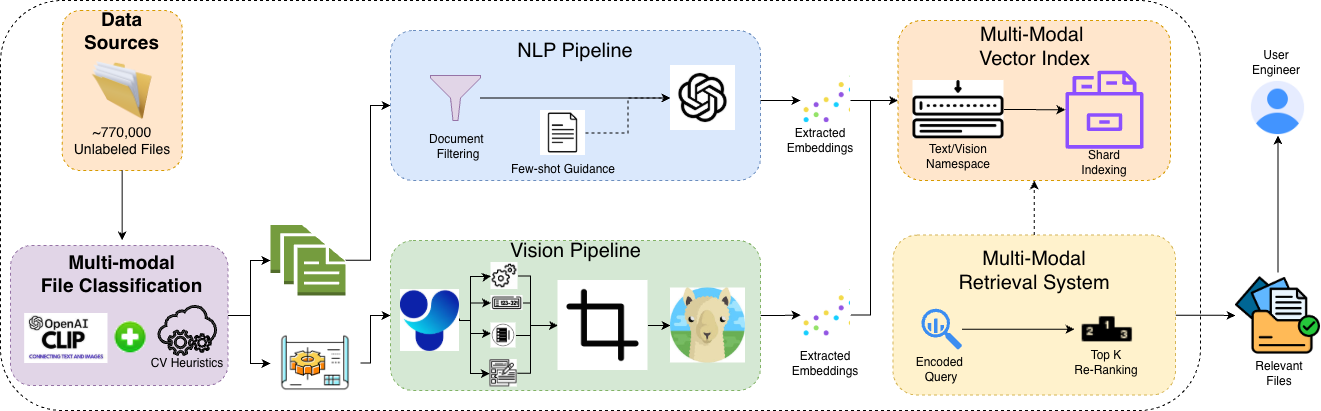}
    \caption{Our multimodal retrieval system processes $\sim$770k unlabeled technical documents through parallel NLP and Vision pipelines. Documents are first routed via zero-shot multimodal classification (CLIP + domain-specific CV heuristics). The NLP pipeline performs document filtering and text embedding extraction, while the Vision pipeline applies preprocessing, spatial cropping, and visual encoding. Both modalities populate a unified multimodal vector index with sharded storage for scalability. At query time, encoded queries retrieve candidates from the index and a re-ranking stage returns the top-$K$ most relevant documents, enabling robust search over heterogeneous document types without manual metadata.}
    \label{fig:arch}
\end{figure*}

\section{\textsc{Blueprint}}
We employ a 2 stage pipeline beginning with document routing (document or drawing), before splitting into an NLP pipeline and Vision pipeline, and combining into a multi-modal retrieval system. Figure \ref{fig:arch} showcases the architecture.

\subsection{Document Routing}
We identify whether a file is an engineering drawing or a document using a zero-shot CLIP classifier \cite{radford2021learningtransferablevisualmodels} augmented with lightweight domain heuristics. Drawings typically contain strong borders ($b^{(i)}$), dense edges ($edge^{(i)}$), and long straight lines ($lines^{(i)}$); these cues form a heuristic score $h^{(i)}$. We additionally include a CAD prior $c^{(i)}$ based on extensions such as \texttt{.dwg} or \texttt{.sldprt}. Thus each file $F^{(i)}$ is represented by $(p^{(i)}_{\text{draw}}, h^{(i)}, c^{(i)})$, where $p^{(i)}_{\text{draw}}$ is the CLIP drawing probability.

To combine these signals, we fit a logistic regression on 500 manually labeled files, modeling $\operatorname{logit} P(y^{(i)}{=}\text{drawing}) = \alpha + \beta_{\text{clip}} p^{(i)}_{\text{draw}} + \beta_{\text{heur}} h^{(i)} + \beta_{\text{cad}} c^{(i)} + \beta_{\text{int}}(p^{(i)}_{\text{draw}} h^{(i)})$. The learned coefficients show that CLIP is the dominant predictor ($\beta_{\text{clip}}\!\approx\!5.91$), while a positive interaction term ($\beta_{\text{int}}\!\approx\!2.21$) indicates that heuristics meaningfully reinforce CLIP when both agree; heuristics alone carry a small negative weight ($\beta_{\text{heur}}\!\approx\!-0.81$).

Empirically, this calibrated combination outperforms both CLIP alone and all heuristic-only baselines: CLIP+heuristics achieves 94.5\% accuracy on the golden test set, improving drawing recall from 94.5\% (CLIP) to 97.7\% while also reducing false positives. This blend of semantic (CLIP) and structural (heuristic) cues yields a reliable routing signal for downstream retrieval.

\subsection{Vision Pipeline}

\textbf{Information Extraction.}
We localize metadata regions with a YOLOv8-S detector~\cite{sohan2024review} trained on our annotations using a multi-task objective (DFL for box regression, CLS, and IoU). On our benchmark it achieves 89.5\% mAP@0.5:0.95 with 5.2\,ms/image on an NVIDIA A100. The detector reliably finds four key regions:\emph{drawing number}, \emph{data block}, \emph{parts list}, and \emph{revisions block}, which anchor downstream extraction.

After localization, we apply a VLM OCR/IE module (``LLaMa-4'') to \emph{each crop} with brief, field-aware prompts. Although the base model is not “lightweight,” the region-only setup (small crops, short prompts, 8-bit quantized inference, capped resolution) yields markedly lower latency than full-page passes while improving exact-field precision. Outputs are normalized and schema-validated before indexing.

\subsection{NLP Pipeline}
\textbf{Document Parsing and Classification.}
After separating drawings from textual documents, we implemented an NLP pipeline to categorize and parse the 392{,}801 document files for downstream normalization and retrieval. Rather than serving as ground-truth labels, these categories provide operational routing for policy- and procedure-focused queries. We used GPT-4o-mini \cite{openai2024gpt4ocard} to assign each document to one of three governance-relevant classes: \emph{policy}, \emph{procedure}, or \emph{other}, via a prompt defining organizational rules (policies), step-wise instructions (procedures), and all remaining technical or administrative documents. Only the first 5 pages (or 8{,}000 characters) were analyzed to balance accuracy with throughput.

Document text was then extracted using a hybrid pipeline: for PDFs, we first applied EasyOCR \cite{easyocr} to rendered pages and fell back to the native text layer (PyMuPDF \cite{mupdf}) when OCR returned no content; DOCX files used python-docx, and other image-based documents were processed directly with EasyOCR. The resulting text was normalized (section headers, step lists, references, units) and embedded using the Nemotron-based textual encoder.

\subsection{Unified Retrieval}\label{sec/retrieval}

\textbf{Shared vector space.}
All documents and drawing regions are embedded into a common vector space \(\mathbf{z}_d \in \mathbb{R}^m\) that fuses text-derived signals (policy/procedure OCR and region-level OCR) with layout-aware features. We use the \textsc{Nemotron-7B} embedding model~\cite{babakhin2025llamaembednemotron8buniversaltextembedding} to obtain a normalized textual embedding \(\mathbf{t}_d\), and we extract a complementary layout/region feature vector \(\mathbf{r}_d\).  
We learn lightweight projection layers \(W_t\) and \(W_r\) and form the final document embedding as \(\mathbf{z}_d = \mathrm{norm}(W_t\,\mathbf{t}_d \oplus W_r\,\mathbf{r}_d)\), where \(\oplus\) denotes concatenation and \(\mathrm{norm}(\cdot)\) is \(\ell_2\) normalization.  
Queries are embedded purely from text using \(\mathbf{z}_q = \mathrm{norm}(W_q\,\mathbf{t}_q)\).

\textbf{Queries}
We evaluate on \(|\mathcal{Q}|=375\) natural-language queries crafted with a domain expert and stratified by intent: \(\mathcal{Q}=\mathcal{Q}_{\text{draw}}\cup\mathcal{Q}_{\text{doc}}\cup\mathcal{Q}_{\text{xmod}}\) with \(|\mathcal{Q}_{\text{draw}}|=150\), \(|\mathcal{Q}_{\text{doc}}|=150\), and \(|\mathcal{Q}_{\text{xmod}}|=75\). These cover drawing-centric lookups (\(\mathcal{Q}_{\text{draw}}\)), policy/procedure lookups (\(\mathcal{Q}_{\text{doc}}\)), and cross-modal queries that should return both drawings and documents (\(\mathcal{Q}_{\text{xmod}}\)).

\paragraph{Query form and normalization.}
Each query \(q\in\mathcal{Q}\) is free text but can be viewed as a slot template \(\phi(q)=\{\texttt{facility},\texttt{asset/part},\texttt{doc\_type},\texttt{constraints}\}\) when applicable (e.g., revision filters, date ranges). We apply light normalization to \(q\) (case folding; punctuation/whitespace canonicalization) prior to retrieval; this only affects matching and does not alter the text presented to judges.

\textbf{Judging protocol.}
For each query \(q\), systems return the top \(k=3\) items.  
Relevance is scored on a 3-point scale \(\ell(q,d)\in\{0,1,2\}\).
To scale evaluation across \(\sim770\mathrm{k}\) documents, we use three independent LLM judges (GPT-5, Llama-3.2-11B-Vision, Mistral-7B), each scoring items in isolation to reduce cross-system bias.
A small, stratified subset is reviewed by three human evaluators to validate agreement and resolve ambiguous cases.

\textbf{Pooling and de-duplication.}
For fair recall estimation, we pool all systems' top-\(k\) results per query, collapse near-duplicates (e.g., successive revisions of the same drawing), and judge the merged set.
Real-world artifacts such as low-quality or partially unreadable scans are kept, while only truly corrupted files are excluded and tracked for coverage.

\textbf{Examples.} Drawing-centric (\(q\in\mathcal{Q}_{\text{draw}}\)):  
\emph{“Facility R8E8700 — voltage monitor on MCC-3; exclude revisions.”}

Policy-centric (\(q\in\mathcal{Q}_{\text{doc}}\)):  
\emph{“Cylinder transport procedure for cryogenic tanks; pre-lift checklist.”}

Multi-modal (\(q\in\mathcal{Q}_{\text{xmod}}\)):  
\emph{“Lockout/tagout policy and corresponding electrical schematic for Main Panel TG:CAB1800.”}

\begin{table*}[h!t]
\centering
\small
\setlength{\tabcolsep}{5pt}
\caption{Overall retrieval performance across 375 benchmark queries (150 vision, 150 NLP, 75 multi-modal).
Each query contributes the top--3 results per system ($\approx$\,6300 document-level judgments).
Reported values are mean~$\pm$~95\%~CI computed over query-level scores.}
\label{tab:overall}
\begin{tabular}{
  l
  S[table-format=1.3(3)]
  S[table-format=1.3(3)]
  S[table-format=1.3(3)]
  S[table-format=1.3(3)]
  S[table-format=1.3(3)]
  S[table-format=1.3(3)]
}
\toprule
System &
\multicolumn{1}{c}{nDCG@3} &
\multicolumn{1}{c}{MAP@3 ($\ge$1)} &
\multicolumn{1}{c}{MAP@3 (=2)} &
\multicolumn{1}{c}{P@3} &
\multicolumn{1}{c}{R@3} &
\multicolumn{1}{c}{Succ@3} \\
\midrule
LLaMA~3.2~Vision        & \num{0.521(0.248)} & \num{0.497(0.244)} & \num{0.330(0.256)} & \num{0.342(0.145)} & \num{0.151(0.052)} & \num{0.623(0.273)} \\
Llama~4~Scout~17B       & \num{0.519(0.158)} & \num{0.503(0.156)} & \num{0.327(0.200)} & \num{0.364(0.096)} & \num{0.137(0.012)} & \num{0.607(0.178)} \\
LLaVA~1.6~Mistral~7B    & \num{0.400(0.246)} & \num{0.378(0.236)} & \num{0.193(0.138)} & \num{0.275(0.154)} & \num{0.109(0.058)} & \num{0.498(0.290)} \\
PaliGemma~2             & \num{0.422(0.240)} & \num{0.395(0.232)} & \num{0.195(0.169)} & \num{0.307(0.166)} & \num{0.122(0.051)} & \num{0.533(0.281)} \\
Pixtral~12B~(2409)      & \num{0.502(0.159)} & \num{0.486(0.153)} & \num{0.314(0.233)} & \num{0.354(0.114)} & \num{0.131(0.022)} & \num{0.592(0.186)} \\
\midrule\midrule
\textbf{\textsc{Blueprint}~(Ours)} &
{\bfseries \num{0.626(0.146)}} &
{\bfseries \num{0.608(0.147)}} &
{\bfseries \num{0.407(0.241)}} &
{\bfseries \num{0.435(0.091)}} &
{\bfseries \num{0.222(0.037)}} &
{\bfseries \num{0.715(0.150)}} \\
\bottomrule
\end{tabular}
\vspace{-4pt}
\end{table*}
\textbf{Scoring and ranking.}
For each document \(d \in C\), we compute a sparse retrieval score \(s_{\text{sparse}}(d \mid q') = \mathrm{BM25}(q', d)\) and a dense retrieval score \(s_{\text{dense}}(d \mid q) = \cos(\mathbf{z}_q, \mathbf{z}_d)\), where \(q'\) is a rewritten or expanded query string used for BM25 and \(q\) is the original text used for dense embeddings. We fuse these with per-query z-score normalization, \(\tilde{s}_{\star}(d) = (s_{\star}(d) - \mu_{\star}(q)) / \sigma_{\star}(q)\), yielding \(s_{\lambda}(d \mid q) = \lambda\,\tilde{s}_{\text{sparse}}(d) + (1 - \lambda)\,\tilde{s}_{\text{dense}}(d)\), where \(\lambda \in [0,1]\) is tuned on a held-out validation set (using a robust variance clamp for small \(|C|\)). A lightweight reranker then promotes region-consistent hits and penalizes revision mismatches: \(s_{\text{final}}(d \mid q) = s_{\lambda}(d \mid q) + \alpha\,\mathrm{match}_{\text{region}}(d,q) + \beta\,\mathrm{consistency}_{\text{rev}}(d,q) - \gamma\,\mathbb{1}[\mathrm{off\text{-}type}(d,q)]\). Here, \(\mathrm{match}_{\text{region}}\in\{0,1\}\) activates when fields extracted from YOLO-detected regions (e.g., title block, parts list, or facility tag) satisfy query constraints, \(\mathrm{consistency}_{\text{rev}}\le 0\) penalizes disallowed revisions, and \(\gamma \ge 0\) optionally penalizes modality mismatches (e.g., drawing vs.\ document) when \textit{Allowed types} are specified. Ties are resolved by normalized document recency (based on date metadata) and page quality (OCR confidence or image resolution). The top-ranked results are returned as \(R_k(q) = \mathrm{TopK}(\{(d, s_{\text{final}}(d \mid q)) : d \in C\}, k = 3)\).

\section{Experiments}
We evaluate \textsc{Blueprint}, our multimodal retrieval framework, on a 5k subset randomly selected from the 558{,}926-document corpus in Section~\ref{sec/dataset}. Our study centers on 370 expert-crafted queries reflecting realistic operational needs. We assess: (i)~\textbf{Retrieval performance} across heterogeneous technical documents (MAP@10, nDCG@10, Precision@10, Recall@10, Success@$\{1,3\}$); (ii)~\textbf{Cross-modal capability} (text→drawing and text→policy, plus mixed intents); and (iii)~\textbf{Component importance} via ablations on routing, extraction quality, and embedding/fusion design. We compare against open-source end-to-end vision–language baselines operating on the same full-page renders.

\subsection{Experimental Setup}
\label{sec:experimental-setup}

\textbf{Metrics.}
We use graded relevance (0/1/2) and report nDCG@3, MAP@3, Precision@3, Recall@3, and Success@$\{1,3\}$. 
We normalize text (case/punctuation), collapse near-duplicate drawing revisions, and measure latency as seconds per file. 
For fair comparison, all systems operate on the same full-page render; VLM baselines score the page directly, while \textsc{Blueprint} applies detect→crop→OCR/IE before indexing. 
Ranking uses a fused score \(s(d\mid q)=\lambda s_{\text{sparse}} + (1-\lambda)s_{\text{dense}}\) with per-query z-normalized dense scores and \(\lambda\) tuned on held-out queries. 
We report 95\% bootstrap CIs, paired randomization tests by query bucket, and expert–LLM agreement (weighted $\kappa$, Kendall’s $\tau$). 
Un-judgeable items are excluded, and coverage is reported.

\textbf{Query set.}
We evaluate on \textbf{375 natural-language queries} crafted with a domain expert: 150 drawing-centric, 150 policy/procedure-centric, and 75 cross-modal (document+drawing). Each query is issued once per system; we collect top-$k$ results with $k=3$ unless stated otherwise. We report metrics per bucket and macro-averaged over all queries.

\textbf{Evaluation protocol.}
To maximize coverage under limited human time and variable scan quality, we use an \emph{LLM-as-judge} protocol as the primary relevance signal. For each query–system pair, the top-$k$ results ($k=3$) are graded on a 0/1/2 rubric over anonymized, randomized items. Near-duplicate pages are pooled and unjudgeable scans excluded. For \emph{human auditing}, three raters label the top-$1$ result on a stratified subset of queries; we report quadratic Cohen’s~$\kappa_w$ and Kendall’s~$\tau$ between expert and LLM nDCG@3, with uncertainty from 10k bootstrap resamples. We additionally report head-to-head \emph{win/lose/tie} rates using nDCG@1 with 95\% CIs.

\textbf{Baselines.}
We compare against end-to-end VLMs that score full-page renders without region crops or OCR pre-processing: \textbf{LLaMA~3.2~Vision} (pooled embedding cosine), \textbf{LLaVA~1.6~Mistral-7B} (embedding head; cross-attention tie-break), \textbf{PaliGemma~2} (pooled VL embeddings), \textbf{Pixtral-12B~(2409)} (pooled embeddings or pairwise scoring), and \textbf{Llama~4~Scout~17B} (pooled similarity; pairwise fallback). We selected baselines that can be run locally to ensure reproducibility, fixed compute budgets, and consistent evaluation across the corpus.

\subsection{Retrieval Performance}\label{sec:retrieval}
We evaluate retrieval quality using both ranking statistics (nDCG, MAP, Precision/Recall, Success) and pairwise win rates.  
All metrics are computed over 375 benchmark queries (150 vision, 150 NLP, and 75 multi-modal), each contributing the top–3 ranked results per system.  
For statistical evaluation, we use three independent LLMs:\textbf{GPT-5}, \textbf{Llama-3.2-Vision}, and \textbf{Mistral-7B}, as judges.  
These models were selected based on their strong performance in the HuggingFace \emph{LLM-as-Judge} leaderboard and were given identical grading prompts.

Table~\ref{tab:overall} reports the mean $\pm$ 95\% confidence intervals of the query-level scores.  
Across all metrics, \textsc{Blueprint} achieves the highest performance, outperforming the next closest baselines (\emph{Llama-4-Scout} and \emph{Llama-3.2-Vision}) by roughly \textbf{10\% absolute} in both Success@3 and nDCG@3.  
These gains indicate that \textsc{Blueprint} not only ranks relevant documents higher but also returns at least one correct result within the top-3 for over 70\% of queries.

Table~\ref{tab:llm-wins} complements these aggregate statistics with \emph{LLM-as-Judge} win-rates in a pairwise “1-vs-1” setting, comparing \textsc{Blueprint} to every other system across 325 queries.  
\textsc{Blueprint} is preferred in \textbf{72.9\%} of head-to-head comparisons on average, while the strongest baseline, \emph{Llama-4-Scout-17B}, attains only \textbf{38.3\%}.  
A sensitivity analysis excluding the weakest baseline (\emph{PaliGemma-2}) yields a consistent win rate of 69.1\%, confirming that the observed improvements are robust across evaluation setups.


\subsection{Human Evaluation}
To validate the reliability of our automated relevance assessments, we conducted a three-person human study consisting of one domain expert and two trained raters. All annotators used the same 3-point relevance scale (0/1/2) as the LLM-based judges and rated the top-1 result for a representative set of 50 queries (25 drawings, 15 NLP, 10 cross-modal). \textbf{Rater agreement.} Human inter-rater reliability is strong: pairwise quadratic Cohen’s~$\kappa$ ranges from 0.49–0.69, and the multi-rater Fleiss’~$\kappa$ is \textbf{0.399}, indicating substantial consistency and establishing a stable human reference for evaluating retrieval quality. \textbf{Human vs.\ LLM agreement.} As shown in Table~\ref{tab:agreement}, expert consensus also exhibits positive agreement with the LLM ensemble used in our automated judging pipeline: across all overlapping items ($n=295$), quadratic $\kappa_w=\textbf{0.31}$ [0.19, 0.42], with comparable values for Drawings (0.35), Procedures (0.25), Cross-Modal (0.25), and Policies (0.24). Although conservative, the LLM judges remain directionally aligned with expert assessments, supporting their use at large scale where human labeling is infeasible. Overall, the human study shows that (i) human raters are internally consistent, (ii) LLM-based 0/1/2 judgments meaningfully track expert consensus, and (iii) \textbf{\textsc{Blueprint}} remains the top-ranked system under both human and automated evaluation.

\begin{table}[h!t]
\centering
\small
\begin{threeparttable}
\caption{LLM-as-judge results showing pairwise wins, losses, and ties against \textsc{Blueprint} across 325 queries (150 vision, 150 NLP, 75 multi-modal). Win~\% is computed as wins/(wins+losses), with ties excluded.}
\label{tab:llm-wins}
\begin{tabular}{lcccr}
\toprule
System & Wins & Losses & Ties & Win \% \\
\midrule
LLaMA~3.2~Vision      & 263 & 645 & 104 & 28.96 \\
Llama~4~Scout~17B     & 331 & 532 & 152 & 38.35 \\
LLaVA~1.6~Mistral~7B  & 166 & 723 & 123 & 18.67 \\
PaliGemma~2           & 117 & 800 & 108 & 12.76 \\
Pixtral~12B           & 328 & 538 & 138 & 37.88 \\
\midrule\midrule
\textbf{\textsc{Blueprint} 17B (overall)} & \textbf{3238} & \textbf{1205} & \textbf{625} & \textbf{72.88}\tnote{\dag} \\
\bottomrule
\end{tabular}
\begin{tablenotes}\footnotesize
\item[\dag] Sensitivity analysis excluding \emph{PaliGemma} (an atypically low baseline) yields a \textsc{Blueprint} win rate of 69.14\%; the main table reports the all-systems result to avoid inflating performance due to a single weak model.
\end{tablenotes}
\end{threeparttable}
\vspace{-4pt}
\end{table}

\begin{table}[h!t]
\centering
\small
\caption{Agreement between expert consensus and the LLM-ensemble on raw 0/1/2 relevance labels. 
Counts ($n$) refer to \emph{items}, each query contributes one top-1 result per system, yielding 295 overlapping human+LLM annotations across the 50 human-evaluated queries. 
$\kappa_w$: quadratic-weighted Cohen’s kappa with bootstrap 95\% CIs.}

\label{tab:agreement}
\begin{tabular}{lccc}
\toprule
Bucket & $n$ & $\kappa_w$ (95\% CI) \\
\midrule
All            & 295 & 0.31 [0.19, 0.42] \\
Drawings       & 149 & 0.35 [0.12, 0.56] \\
Procedures     &  51 & 0.25 [0.07, 0.43] \\
Multi-Modal    &  59 & 0.25 [0.08, 0.44] \\
Policies       &  24 & 0.24 [-0.01, 0.51] \\
\bottomrule
\end{tabular}
\vspace{-4pt}
\end{table}

\begin{table*}[t]
\centering
\small
\setlength{\tabcolsep}{5pt}
\caption{OCR ablations on \textbf{val/test} with LLM-as-judge (\textit{GPT-5}, \textit{LLaMA-3}, \textit{Phi-3.5-mini}). We report top-3 retrieval metrics (nDCG@3, MAP@3, P@3, R@3, Succ@3) as \emph{means across judges} (per-judge CIs in the appendix). All variants share identical indexing and reranking; only the OCR stage and region regime differ (traditional OCR vs.\ \textsc{Blueprint} VLM-OCR, predicted boxes, oracle GT boxes, or full-page).}

\label{tab:ablation_ocr_gpt_top3}
\begin{tabular}{
  l
  S[table-format=1.3]
  S[table-format=1.3]
  S[table-format=1.3]
  S[table-format=1.3]
  S[table-format=1.3]
  S[table-format=1.3]
}
\toprule
\textbf{System Variant} &
\multicolumn{1}{c}{\textbf{nDCG@3}} &
\multicolumn{1}{c}{\textbf{MAP@3 ($\ge$1)}} &
\multicolumn{1}{c}{\textbf{MAP@3 (=2)}} &
\multicolumn{1}{c}{\textbf{P@3}} &
\multicolumn{1}{c}{\textbf{R@3}} &
\multicolumn{1}{c}{\textbf{Succ@3}} \\
\midrule
Tesseract (Pred boxes)            & \num{0.532} & \num{0.525} & \num{0.427} & \num{0.353} & \num{0.582} & \num{0.582} \\
Tesseract (\textbf{Oracle boxes}) & \num{0.457} & \num{0.439} & \num{0.384} & \num{0.261} & \num{0.509} & \num{0.509} \\
\hline
EasyOCR (Pred boxes)              & \num{0.525} & \num{0.513} & \num{0.440} & \num{0.323} & \num{0.573} & \num{0.573} \\
EasyOCR (\textbf{Oracle boxes})   & \num{0.390} & \num{0.373} & \num{0.304} & \num{0.235} & \num{0.442} & \num{0.442} \\
\hline
Tesseract (Fullpage) \emph{diag.} & \num{0.208} & \num{0.197} & \num{0.128} & \num{0.108} & \num{0.240} & \num{0.240} \\
EasyOCR (Fullpage) \emph{diag.}   & \num{0.287} & \num{0.270} & \num{0.181} & \num{0.159} & \num{0.338} & \num{0.338} \\
\hline
\textsc{\textbf{Blueprint}} (Pred boxes, \textbf{VLM-OCR}) &
\multicolumn{1}{c}{\textbf{\num{0.563}}} &
\multicolumn{1}{c}{\textbf{\num{0.539}}} &
\multicolumn{1}{c}{\textbf{\num{0.458}}} &
\multicolumn{1}{c}{\textbf{\num{0.313}}} &
\multicolumn{1}{c}{\textbf{\num{0.644}}} &
\multicolumn{1}{c}{\textbf{\num{0.644}}} \\
\textsc{\textbf{Blueprint}} (\textbf{Oracle boxes}, \textbf{VLM-OCR}) &
\multicolumn{1}{c}{\textbf{\num{0.699}}} &
\multicolumn{1}{c}{\textbf{\num{0.675}}} &
\multicolumn{1}{c}{\textbf{\num{0.601}}} &
\multicolumn{1}{c}{\textbf{\num{0.401}}} &
\multicolumn{1}{c}{\textbf{\num{0.780}}} &
\multicolumn{1}{c}{\textbf{\num{0.780}}} \\
\bottomrule
\end{tabular}
\vspace{-4pt}
\end{table*}

\subsection{Ablation Studies}
\textbf{OCR Ablations.}
\label{sec:ocr-ablation}
We ablate only the OCR stage while holding indexing, embeddings, reranking, and normalization fixed (Table~\ref{tab:ablation_ocr_gpt_top3}). We compare Tesseract \cite{kay2007tesseract} and EasyOCR  \cite{easyocr} with (i) \emph{predicted} \textsc{Blueprint} boxes, (ii) \emph{oracle} (GT) boxes, and (iii) a \emph{full-page} diagnostic; we also evaluate our \textsc{Blueprint} VLM-OCR under the same box regimes. Relevance is judged by an LLM ensemble (GPT-5, LLaMA~3, Phi-3.5) and reported as means averaged across judges (per-judge 95\% CIs in the appendix).

Three takeaways emerge. \textbf{(1) Layout-aware OCR beats full-page OCR:} both Tesseract and EasyOCR drop substantially on full-page runs. \textbf{(2) Predicted boxes are competitive with oracle boxes for classic OCR,} indicating our detector reliably localizes informative regions. \textbf{(3) \textsc{Blueprint} VLM-OCR is strongest overall:} with oracle boxes it achieves \textbf{0.699} nDCG@3, \textbf{0.675} MAP@3 ($\ge$1), and \textbf{0.780} Succ@3; with predicted boxes it remains ahead of traditional engines (nDCG@3 \textbf{0.563}, Succ@3 \textbf{0.644}). These results underscore that accurate region proposals \emph{and} a multimodal, drawing-aware OCR are both critical for retrieval in engineering archives.


\begin{figure}[h!t]
    \centering
    \includegraphics[width=.45\textwidth]{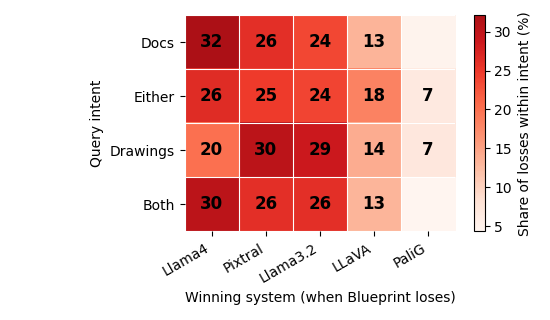}
    \caption{Heatmap showing which baselines win against \textsc{Blueprint} across query intents. “Either” entries reflect queries where judges accept drawings or documents, and values show the percentage of losses per intent.}

    \label{fig:error_heatmap}
\end{figure}

\subsection{Error analysis}
Across all pairwise comparisons where \textsc{Blueprint} loses under the Llama-3 judge, the loss rate is highest for document-only queries (DOCS, 35.9\%) and auto-generated queries (AUTO, 27.5\%), and markedly lower for drawing-centric intents (DRAWINGS, 20.1\%; BOTH, 19.2\%). This matches our design goal: \textsc{Blueprint} is specialized for engineering drawings and trades some performance on generic document retrieval. When it does lose, the winners are typically large full-page VLMs (e.g., Llama-4-Scout-17B, Pixtral-12B), with smaller models rarely outperforming it. Qualitative inspection highlights three recurring failure modes: modality routing errors (retrieving drawings instead of procedures), region/OCR misses on fine-grained attributes, and cross-document reasoning failures, which suggest future improvements in document routing, region-level OCR, and document–drawing linkage. Figure \ref{fig:error_heatmap} displays a heatmap of loss percantage broken down by query type.

\begin{figure}[h!t]
    \centering
    \includegraphics[width=.45\textwidth]{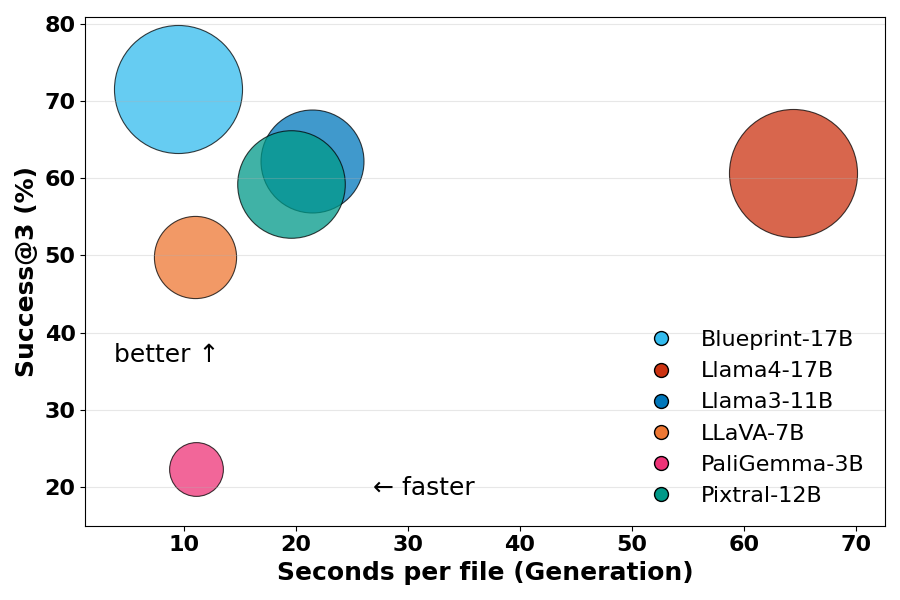}
    \caption{Latency–accuracy trade-off across systems. \textsc{Blueprint} attains the fastest end-to-end processing time ($\approx 9.46$\,s/file) and the highest retrieval accuracy ($\mathrm{Succ}@3 = 71.5\%$), outperforming all vision–language baselines. Bubble size indicates parameter count.
}

    \label{fig:fps_graph}
\end{figure}

\section{Discussion}

\textbf{Key Insights.}
\textsc{Blueprint} delivers the best combination of accuracy, reliability, and efficiency across our full 770k-document environment. Against the strongest baseline (Llama-4-Scout-17B), paired bootstrap tests show large, statistically significant improvements for engineering-drawing retrieval (e.g., nDCG@3: $\Delta=+0.26$, 95\% CI [0.22, 0.31]; Succ@3: $\Delta=+0.27$, CI [0.22, 0.32]). For multimodal queries the systems are statistically tied, and for document-only queries Scout shows a small, non-significant advantage.

Despite using a multi-stage pipeline, \textsc{Blueprint} is the fastest system, achieving 9.46 s/file versus 64.36 s/file for Llama-4-Scout and 19.6–21.5 s/file for other VLMs, as seen in Figure \ref{fig:fps_graph}. It also achieves the highest Success@3 (71.5\%), substantially above the strongest baseline (60.7\%). Overall, \textsc{Blueprint} provides both statistical gains and operational advantages, especially for drawing-centric search scenarios that dominate real engineering workflows.

\textbf{Limitations.}
Our metadata schema was chosen with expert engineers as the most critical fields for retrieval/compliance, but it is not exhaustive; site-specific cues (e.g., vendor callouts, atypical “as-built” notes) may be missed. We also assume readable scans and identifiable regions, so severe degradation or nonstandard title blocks can reduce fidelity. The reranker currently handles a small set of constraints (facility/tag/revision) and does not yet model richer cross-document relations.



\section{Conclusions}
We presented \textsc{Blueprint}, a layout-aware multimodal retrieval framework for engineering documents and drawings. By coupling region detection, VLM-based OCR, and normalized text in a unified embedding space, \textsc{Blueprint} delivers state-of-the-art performance on 375 graded queries (e.g., \textbf{0.626} nDCG@3, \textbf{0.715}~Succ@3), outperforming strong open-source VLM baselines. The results highlight the value of structured, region-aware extraction for technical corpora.

The approach generalizes to diagram-heavy domains such as BIM/architecture, aerospace schematics, and technical manuals, enabling applications in cross-modal search, compliance, maintenance, and archive modernization.

\textbf{Future Work.} We will extend extraction to relation graphs, introduce lightweight rerankers for hard negatives, and further improve robustness to degraded scans and multilingual symbols. In parallel, we plan to release a cross-domain benchmark and organize a community shared task.

{
    \small
    \bibliographystyle{ieeenat_fullname}
    \bibliography{main}
}

\newpage
\appendix
\section{Additional Ablations}
\subsection{Zero-Shot Document Classification}
The first step in our framework was to distinguish between engineering drawings and textual documents. We evaluated four approaches on the withheld golden test set of 1{,}500 files. Baseline methods performed poorly: the NLP-based \texttt{BART-large} model achieved only 49.1\% accuracy overall, with near-zero recall on drawings (0.8\%) and a corresponding F1 score of just 1.5\%. A heuristics-based baseline, implemented as a logistic regression classifier over structural features (e.g., edges, lines, shapes), performed similarly with 48.4\% accuracy, achieving moderate performance on drawings (F1 = 0.61) but very low performance on documents (F1 = 0.24).

\begin{figure}[th!]
    \centering
    \includegraphics[width=0.95\columnwidth]{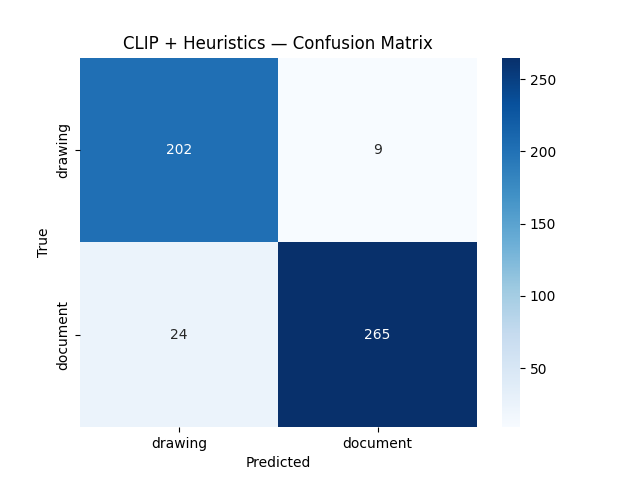}
    \caption{Confusion matrix for the drawing vs.\ document classification task. 
    Values are normalized by row. (Clip + Heuristics)}
    \label{fig:confusion-matrix-apdx}
\end{figure}

In contrast, vision-based methods provided a substantial improvement. CLIP achieved 93.3\% accuracy, with balanced performance across both categories (F1 = 0.93 for documents and 0.93 for drawings). When combined with heuristics, CLIP’s performance further improved, reaching 94.5\% overall accuracy, with F1 scores of 0.94 for documents and 0.95 for drawings, the best results across all approaches. These results confirm that multimodal, vision-driven zero-shot methods are highly effective for large-scale filtering of engineering archives, outperforming both NLP and heuristic baselines by a wide margin (Table~\ref{tab:model-compare}, Figure~\ref{fig:confusion-matrix-apdx}).

\begin{table}[th!] 
\centering
\small
\caption{Zero-shot performance (drawing vs.\ document) on the withheld golden test set (1500 files). Best values per column are in \textbf{bold}.}
\label{tab:model-compare}
\setlength{\tabcolsep}{3pt}
\begin{tabularx}{\columnwidth}{ll|@{\hskip 4pt}YYYY}
\toprule
Model & Type &  Prec. & Rec. & F1 & Acc. \\
\midrule
\multirow{2}{*}{Heuristics only} 
    & Drawing  &  0.490 & 0.808 & 0.610 & \multirow{2}{*}{0.484} \\
    & Document &  0.455 & 0.160 & 0.237 &  \\
\midrule
\multirow{2}{*}{BART-large}      
    & Drawing  &  0.240 & 0.008 & 0.015 & \multirow{2}{*}{0.491} \\
    & Document &  0.496 & \textbf{0.975} & 0.657 &  \\
\midrule
\multirow{2}{*}{CLIP} 
    & Drawing  &  \textbf{0.922} & 0.945 & 0.934 & \multirow{2}{*}{0.933} \\
    & Document &  0.944 & 0.920 & 0.932 &  \\
\midrule
\multicolumn{6}{c}{} \\[-6pt] 
\midrule
\multirow{2}{*}{\textbf{CLIP+Heuristics}} 
    & Drawing  &  0.919 & \textbf{0.977} & \textbf{0.947} & \multirow{2}{*}{\textbf{0.945}} \\
    & Document &  \textbf{0.976} & 0.913 & \textbf{0.944} &  \\
\bottomrule
\end{tabularx}
\end{table}

\subsection{Information Detection}
Accurate region detection is a prerequisite for all downstream OCR and identifier
extraction. We therefore evaluated a range of modern object-detection models on
our engineering-drawing detection benchmark. Our primary detector is YOLOv8-S,
trained on 500 annotated drawings and evaluated on a held-out set of 250 unseen
drawings. We use YOLOv8-S because it provides an excellent accuracy–latency
trade-off: as shown in Table~\ref{tab:vision_ablation}, YOLOv8-S matches the
accuracy of newer variants such as YOLOv11-S while running substantially faster.

To assess robustness, we ablate YOLOv8-S against (i)~other YOLO families (YOLOv5-S, YOLOv10-S, YOLOv11-S) and (ii)~state-of-the-art transformer-based detectors (RT-DETR, Faster~R-CNN, RetinaNet). YOLOv8-S achieves the highest mAP@0.5:0.95 among all real-time models, with competitive precision/recall and very low latency, making it well suited for high-volume archival processing where millions of pages must be scanned quickly. Reliable detection ensures that subsequent OCR operates on clean, semantically meaningful regions (e.g., title blocks, revision tables, parts lists), which directly improves identifier normalization and downstream retrieval accuracy.

\begin{table*}[t]
\centering
\caption{Comparison of detection models on the withheld engineering drawing test dataset. 
All accuracy values are percentages; latency measured in milliseconds per image.}
\begin{tabular}{lcccccc}
\toprule
\textbf{Model} & \textbf{mAP@0.5:0.95} & \textbf{Recall} & \textbf{Precision} & \textbf{Params (M)} & \textbf{GFLOPs} & \textbf{Latency (ms)} \\
\midrule
YOLOv5-S & 87.94 & 90.56 & \textbf{97.36} & 9.12 & 24.05 & \textbf{2.68} \\
\textbf{YOLOv8-S} & \textbf{89.50} & 92.61 & 96.51 & 11.14 & 28.66 & 5.16 \\
YOLOv10-S & 85.17 & 88.89 & 93.54 & 8.07 & 24.79 & 92.23 \\
YOLOv11-S & \textbf{89.50} & 92.13 & 95.83 & 9.43 & 21.56 & 81.37 \\
\midrule
Faster R-CNN (R-50 FPN) & 78.11 & 90.88 & 89.69 & 41.0 & 180.3 & 45.10 \\
RetinaNet (R-50 FPN) & 55.55 & 90.58 & 79.43 & 36.4 & 150.2 & 33.00\\
RT-DETR-L & 88.51 & \textbf{93.28} & 96.95 & 32.82 & 108.0 & 27.87 \\
\bottomrule
\end{tabular}
\label{tab:vision_ablation}
\end{table*}

\section{Evaluation Metrics}
For clarity and reproducibility we define all metrics in this section.
\subsection{Detection \& Classification Metrics}
We evaluated the performance of our vision pipeline against a manually annotated subset of the corpus using standard classification metrics.

\paragraph{Accuracy.}  
Overall accuracy measures the proportion of correctly classified files:
\begin{equation}
    \text{Acc} = \frac{1}{N} \sum_{i=1}^{N} \mathbf{1}\{\hat{y}^{(i)} = y^{(i)}\},
\end{equation}
where $y^{(i)}$ is the ground-truth label for file $i$, $\hat{y}^{(i)}$ is the predicted label, and $\mathbf{1}\{\cdot\}$ is the indicator function.

\paragraph{Precision and Recall.}  
For each class $k \in \{\text{drawing}, \text{document}\}$, we define true positives (TP$_k$), false positives (FP$_k$), and false negatives (FN$_k$). Precision and recall are then:
\begin{align}
    \text{Prec}_k &= \frac{\text{TP}_k}{\text{TP}_k + \text{FP}_k}, \\
    \text{Rec}_k  &= \frac{\text{TP}_k}{\text{TP}_k + \text{FN}_k}.
\end{align}

\paragraph{F1 Score.}  
The F1 score is the harmonic mean of precision and recall:
\begin{equation}
    F1_k = \frac{2 \cdot \text{Prec}_k \cdot \text{Rec}_k}{\text{Prec}_k + \text{Rec}_k}.
\end{equation}

\paragraph{Macro Average.}  
To summarize performance across both classes, we report the macro-average of each metric:
\begin{equation}
    \text{Macro-}M = \frac{1}{2}\sum_{k \in \{\text{drawing}, \text{document}\}} M_k,
\end{equation}
where $M_k$ denotes precision, recall, or F1 for class $k$.

\subsection{Retrieval Metrics}
For the end-to-end retrieval benchmark (Table~\ref{tab:overall}), we evaluate each
system using standard information–retrieval metrics adapted to our graded
relevance scheme (0/1/2). Because many computer vision readers may be less
familiar with these metrics, we briefly define them here.

\paragraph{nDCG@3.}
Normalized Discounted Cumulative Gain measures how well a ranked list places
highly relevant items near the top. For a query $q$:
\begin{equation}
    \text{DCG@3}(q) = \sum_{i=1}^{3} \frac{\ell(q,d_i)}{\log_2(i+1)},
\end{equation}
\begin{equation}
    \text{nDCG@3}(q) = \frac{\text{DCG@3}(q)}{\text{IDCG@3}(q)},
\end{equation}
where IDCG@3 is the maximum achievable DCG for that query.

\paragraph{MAP@3.}
Mean Average Precision evaluates how consistently relevant items appear across the
top--3 positions:
\begin{equation}
    \text{AP@3}(q)
    = \frac{1}{R(q)}
      \sum_{i=1}^{3} \text{Prec@}i(q)\,
      \mathbf{1}\bigl\{\ell(q,d_i) > 0 \bigr\},
\end{equation}
where $R(q)$ is the number of relevant items.  
We report two versions:
\begin{itemize}
    \item \textbf{MAP@3($\geq$1)} — counts any relevance $\ell>0$,
    \item \textbf{MAP@3(=2)} — requires the highest relevance grade.
\end{itemize}

\paragraph{Precision@3 and Recall@3.}
Precision@3 measures the fraction of top--3 items that are relevant; Recall@3 measures
the fraction of all relevant items retrieved:
\begin{equation}
    \text{P@3}(q) = \frac{1}{3}\sum_{i=1}^{3}
        \mathbf{1}\bigl\{\ell(q,d_i) > 0\bigr\},
\end{equation}
\begin{equation}
    \text{R@3}(q) = \frac{1}{R(q)}\sum_{i=1}^{3}
        \mathbf{1}\bigl\{\ell(q,d_i) > 0\bigr\}.
\end{equation}

\paragraph{Success@3.}
Success@3 indicates whether at least one relevant item appears in the top--3:
\begin{equation}
    \text{Succ@3}(q)
    = \mathbf{1}\!\left\{
        \max_{i \le 3} \ell(q,d_i) \ge 1
      \right\}.
\end{equation}

\paragraph{Aggregation.}
All metrics are computed per query and then averaged across the 375 benchmark
queries (150 vision-only, 150 NLP-only, 75 multimodal). We report
mean~$\pm$~95\% confidence intervals over the per-query scores.

\section{Baseline Models and Prompting Protocols}
To ensure fair and reproducible comparison across systems, we document all
baseline models, prompts, and decoding settings used in the retrieval
experiments. All systems—including \textsc{Blueprint}—were evaluated under the
same input constraints and returned their top--$k$ ranked items for every query.

\subsection{A. Model Suite}
We evaluate six state-of-the-art vision–language models (VLMs) alongside our
proposed \textsc{Blueprint} system:

\begin{itemize}
    \item \textbf{LLaMA~3.2~Vision~11B} (full-page vision encoder + LLaMA-3.2 decoder)
    \item \textbf{Llama-4-Scout~17B} (OCR-heavy retrieval-oriented model)
    \item \textbf{LLaVA-1.6-Mistral~7B}
    \item \textbf{PaLI-Gemma~2~3B}
    \item \textbf{Pixtral~12B}
    \item \textbf{Phi-3.5-Vision} (ablations only; struggled with full-page inputs)
    \item \textbf{\textsc{Blueprint}} (ours: region-aware pipeline + hybrid retrieval)
\end{itemize}

All models were run exactly as released by their maintainers, with no fine-tuning,
task-specific adapters, or retrieval-engine modifications. Models were limited to
the same maximum input resolution and were given access only to full-page image
renders (no region hints, no bounding boxes, and no OCR text unless produced by
the model itself).

\subsection{Prompting Protocol}
To avoid inadvertently favoring any model family, all baselines received the same
natural-language retrieval prompt for each query, presented in Figure \ref{fig:model_prompts}.

\begin{figure*}[t]
\centering
\scriptsize
\begin{minipage}{0.97\textwidth}
\begin{verbatim}
You are an ingestion agent for engineering archives. You will be shown ONE page as an image.
Your job has THREE steps:

(1) CLASSIFY
Decide the document type. Choose exactly ONE from:
- ENGINEERING_DRAWING
- POLICY
- PROCEDURE
- OTHER

(2) EXTRACT
Depending on the type, extract strongly structured content:

A. If it is an ENGINEERING_DRAWING:
  - drawing_number: the main drawing / print / figure / doc number in the title block
  - title_block_text: all text in the title block
  - revision_block_text: all text in the revision/change block
  - parts_list_or_bom: any tabular/parts list, row by row
  - notes: general notes, welding/material notes, callouts
  - ALSO output the full plain-text reading of the entire page

B. If it is a POLICY:
  - policy_id or number if present
  - title / subject
  - section_headings (in order)
  - body_text (full text)
  - tables (if visible)

C. If it is a PROCEDURE:
  - procedure_id or number if present
  - title
  - steps (numbered or bulleted) in order
  - prerequisites / scope / purpose if visible
  - tables / forms if visible
  - body_text (full text)

D. If it is OTHER:
  - title or top heading if any
  - body_text (full text)
  - tables if visible

(3) OUTPUT
Return ONLY JSON in this shape:

{
  "doc_type": "<ENGINEERING_DRAWING | POLICY | PROCEDURE | OTHER>",
  "text": "<full-page plain text with line breaks>",
  "drawing_fields": {
    "drawing_number": "...",
    "title_block_text": "...",
    "revision_block_text": "...",
    "parts_list_or_bom": "...",
    "notes": "..."
  },
  "policy_fields": {
    "policy_id": "...",
    "title": "...",
    "section_headings": "...",
    "tables": "..."
  },
  "procedure_fields": {
    "procedure_id": "...",
    "title": "...",
    "steps": "...",
    "tables": "..."
  }
}
\end{verbatim}
\end{minipage}
\caption{Prompt used for full-page ingestion and structured extraction for all baseline models.}
\label{fig:model_prompts}
\end{figure*}

No system was allowed to “cheat” by inspecting folder paths, filenames outside the
render, or metadata unavailable to others.

\paragraph{Constraint handling.}
Baselines were not provided any additional engineering-domain heuristics. For
example, a request for “drawings size~E” must be inferred purely from OCR or visual
inspection, not any model-specific rules. This ensured that improvements arise from
\textsc{Blueprint}'s structured extraction and not from query leakage or handcrafted
logic.

\subsection{Decoding and Search Parameters}
All systems used deterministic decoding:

\begin{itemize}
    \item \texttt{do\_sample = False} (greedy decoding; no sampling),
    \item \texttt{top\_p = 1.0} (default, unused when \texttt{do\_sample = False}),
    \item \texttt{max\_new\_tokens = 512} per page.
\end{itemize}

\paragraph{Image preprocessing.}
All baselines received:
\begin{itemize}
    \item full-page (PNG/PDF) renders at fixed resolution (2048 px max side),
    \item no region crops,
    \item no pre-extracted title blocks, revision tables, or parts lists.
\end{itemize}

This ensured that any advantage from structured region extraction originates
from \textsc{Blueprint} itself.

\subsection{Blueprint Query Pipeline}
For completeness, \textsc{Blueprint} received the same text query but used its
internal multimodal pipeline:

\begin{enumerate}
    \item zero-shot document/drawing classification,
    \item YOLOv8-S region detection,
    \item VLM-based OCR for title blocks, parts lists, revision tables, etc.,
    \item identifier normalization (DWG number, rev, facility code),
    \item hybrid sparse+dense retrieval with region-aware reranking.
\end{enumerate}

\subsection{Fairness Considerations}
To ensure no model was advantaged:
\begin{itemize}
    \item All models saw the same query wording.
    \item All models received identical image renders.
    \item No model received privileged internal metadata.
    \item All were capped at three final ranked outputs.
    \item All judgments (human+LLM) were done on anonymized outputs.
\end{itemize}

This evaluation design emphasizes retrieval ability, not language-model fluency or
caption length.

\section{Human Evaluation}

\textbf{System rankings.}  
Each system’s relevance score was averaged across human raters, and per-query nDCG@1 was computed with bootstrap 95\% confidence intervals. Table~\ref{tab:human_eval_succ1} shows that \textsc{Blueprint} achieves the highest human-validated retrieval accuracy, with mean nDCG@1 = \textbf{0.74}~$\pm$~0.12.  
The strongest baseline, Llama-3.2-11B-Vision, reaches 0.49~$\pm$~0.14, followed by Llama-4-Scout-17B (0.43~$\pm$~0.14).  

Full-page VLMs such as Pixtral-12B (0.35~$\pm$~0.14) and PaLI-Gemma-2-3B (0.18~$\pm$~0.11) underperform substantially, highlighting that generic vision–language reasoning does not transfer well to engineering diagrams or multi-region technical documents.  
Blueprint’s margin is widest on multi-modal queries (1.00~$\pm$~0.00), where accurate integration of layout, identifiers, and text is essential.

\begin{table}[t]
\centering
\scriptsize
\setlength{\tabcolsep}{4pt}
\caption{Human evaluation success at $top_k{=}1$ (Succ@1): mean~$\pm$~95\%~CI over queries. Three raters (1 expert, 2 trained raters). 50 queries: 25 vision, 15 NLP, 10 multi-modal.}
\label{tab:human_eval_succ1}
\begin{tabular}{l
                S[table-format=1.2(2)]
                S[table-format=1.2(2)]
                S[table-format=1.2(2)]
                S[table-format=1.2(2)]}
\toprule
\textbf{System} & \multicolumn{1}{c}{All} & \multicolumn{1}{c}{Vision} & \multicolumn{1}{c}{NLP} & \multicolumn{1}{c}{MM} \\
\midrule
Llama-3.2-11B-V      & 0.49 \pm 0.14 & 0.33 \pm 0.19 & 0.73 \pm 0.20 & 0.50 \pm 0.30 \\
L4-Scout-17B         & 0.43 \pm 0.14 & 0.36 \pm 0.18 & 0.29 \pm 0.21 & 0.80 \pm 0.25 \\
LLaVA-1.6-M          & 0.31 \pm 0.12 & 0.36 \pm 0.18 & 0.29 \pm 0.21 & 0.20 \pm 0.25 \\
PaLI-Gemma-2-3B      & 0.18 \pm 0.11 & 0.24 \pm 0.16 & 0.13 \pm 0.17 & 0.10 \pm 0.15 \\
Pixtral-12B          & 0.35 \pm 0.14 & 0.24 \pm 0.16 & 0.57 \pm 0.29 & 0.33 \pm 0.33 \\
\midrule\midrule
\textbf{Blueprint}   & \textbf{0.74 $\pm$ 0.12} & \textbf{0.72 $\pm$ 0.18} & \textbf{0.60 $\pm$ 0.27} & \textbf{1.00 $\pm$ 0.00} \\
\bottomrule
\end{tabular}
\end{table}

\subsection{Inter-Rater Reliability}
To quantify the consistency of human annotations, we measured pairwise and
multi-rater agreement across the three evaluators (1 expert, 2 trained raters)
using the same graded relevance labels (0/1/2) used throughout the benchmark.

\paragraph{Pairwise agreement.}
Across the overlapping items for each pair of raters, we observe strong
alignment:

\begin{itemize}
    \item \textbf{rater 1 vs.\ rater 2:} 324 overlapping items, 72.8\% raw agreement, 
    Cohen’s $\kappa{=}0.36$ (unweighted), and quadratic-weighted 
    $\kappa_w{=}0.52$.

    \item \textbf{rater 1 vs.\ rater 3:} 300 overlapping items, 76.7\% raw agreement, 
    Cohen’s $\kappa{=}0.51$, and $\kappa_w{=}0.69$.

    \item \textbf{rater 2 vs.\ rater 3:} 324 overlapping items, 77.2\% raw agreement,
    Cohen’s $\kappa{=}0.36$, and $\kappa_w{=}0.49$.
\end{itemize}

Quadratic-weighted kappas between 0.49–0.69 indicate \emph{moderate to strong}
agreement for a 3-level ordinal scale, consistent with expectations for 
fine-grained 0/1/2 relevance judgments over heterogeneous engineering documents.

\paragraph{Model-level consistency.}
Per-variant agreement is generally high across raters (typically 0.74–0.89).
\textsc{Blueprint} shows somewhat lower pairwise agreement (0.42–0.63), not due
to inconsistency, but because it more frequently surfaces borderline-relevant
cases (e.g., multiple contextually plausible drawings), which require careful
disambiguation. This is consistent with the system achieving the highest
recall-oriented and multi-modal retrieval performance.

\paragraph{Multi-rater reliability.}
Considering all three raters simultaneously, 290 items were annotated by all
raters. Full agreement occurred on 63.8\% of items, and Fleiss’ 
$\kappa{=}0.40$, again indicating moderate agreement across the 0/1/2
relevance scale.

\medskip
Together, these results demonstrate that the human annotations used to validate
\textsc{Blueprint} are consistent and reliable across raters, and that the
graded relevance scheme produces stable judgments despite the diversity of
technical documents involved.

\section{LLM as Judge}
To evaluate models and reduce human workload in a real-world engineering setting, we implement a framework for ``LLM as judge'' in both an arena setting and a statistical analysis. These judges and statistics are discussed in Section~\ref{sec:retrieval} of the main paper.  

\textbf{Decoding settings.}
All LLM judges used fully deterministic decoding to ensure strict reproducibility.
We set: temperature $=0$, top\_p $=1$, and max\_tokens $=4000$ for Arena judgments
(with effectively unlimited tokens for the JSON scoring task). No sampling, nucleus
randomness, or system-level stochasticity was used beyond the randomized
A/B model order. This guarantees that all variation in judgments is due solely
to model behavior, not decoding noise

\textbf{Why LLM as Judge?}
Engineering archives at academic and industrial facilities contain hundreds of thousands of unlabeled files. Full human judgment is infeasible at scale: manuallyLabeling the 375-query × 6-model matrix would require \(\sim 6{,}300\) 
individual relevance annotations, which is infeasible to perform manually. ``LLM as judge'' reduces this cost substantially, while maintaining human-level reliability through a hybrid LLM–human framework.

\textbf{Arena.} Three judges (GPT-5, Llama-3.2, Mistral-7B; for ablations, Phi-3.5 replaced Mistral-7B but struggled on full-page prompts). To ensure fairness, the order of systems was randomized: \textsc{Blueprint} appeared as ``System~A'' exactly 
50\% of the time and ``System~B'' the other 50\%. This mitigates positional bias, where a judge, if uncertain, may default to the first system. All judges received identical prompts and decoding settings.

Figure~\ref{fig:llm-judge-arena-prompt} shows the exact prompt used in the arena 
evaluation, while Table~\ref{tab:arena_examples_sizeE} presents GPT-5’s judgments 
for the query $q = \text{``drawings size E and only 1 sheet''}$.

\begin{figure*}[t]
\centering
\begin{minipage}{0.95\textwidth}
\footnotesize
\begin{verbatim}
You are an objective judge for retrieval from a legacy engineering archive.

You will receive:
- A user query Q. Q may target drawings, documents, or BOTH (multimodal).
- Two anonymous systems (System A and System B). Each system returns up to 5 ranked items.
- Each returned item may be a DRAWING or a DOCUMENT; items include minimal metadata
  (e.g., drawing number, sheet, rev, material, facility/building, parts count,
  doc id/code, title, author, date).

Your task:
Choose which system is better for this query: A, B, or tie.

Judging rules (apply in order):
A. INTENT MATCH
   - If Q names an identifier (e.g., drawing "drawing num", code xyz),
     systems that return the exact target at rank 1 (or near the top) should be preferred.
   - If Q requires BOTH (asks for a procedure/policy/document AND a drawing),
     prefer systems whose top results collectively satisfy BOTH in the top-5.
   - If Allowed types are provided, treat them as a hard constraint.

B. CONSTRAINT SATISFACTION
   - Prefer items that satisfy secondary constraints (building,
     material, rev, sheet, parts count, facility, date, author, vendor).

C. RANKING QUALITY
   - Correct items appearing higher are better than the same items lower.
   - Penalize unrelated items, hallucinated IDs, or near-duplicate items.

D. MULTIMODAL PAIRING (when Q is BOTH)
   - Extra credit when both modalities appear with consistent anchors.
   - Partial satisfaction is strictly worse than full satisfaction.

Tie-breaking (only if needed):
1) More exact anchor matches.
2) More constraints satisfied.
3) Higher useful density in ranks 1--2.
4) Fewer unrelated items.

Output STRICTLY JSON ONLY:
{
  "winner": "A" | "B" | "tie",
  "explanation": "brief reason (1-2 sentences)"
}
\end{verbatim}
\end{minipage}
\caption{Arena-style prompt used for the LLM-as-judge comparison between two anonymous retrieval systems. All judges and models received the identical prompt; only the system outputs were permuted.}
\label{fig:llm-judge-arena-prompt}
\end{figure*}

\begin{table*}[t]
\centering
\footnotesize

\begin{center}
\begin{tabular}{c}
\toprule
\textbf{Query:} ``drawings size E and only 1 sheet'' \\
\bottomrule
\end{tabular}
\end{center}


\begin{tabular}{p{0.20\linewidth} p{0.20\linewidth} p{0.06\linewidth} p{0.44\linewidth}}
\toprule
\textbf{Model A} & \textbf{Model B} & \textbf{Winner} & \textbf{Judge Explanation} \\
\midrule

Blueprint & Llama-3.2-Vision  
& A  
& Blueprint returns engineering drawings matching \emph{SIZE E} and \emph{1 of 1 sheets}, while B returns off-type documents. \\[6pt]

llava-v1.6-mistral-7b-hf & Blueprint  
& B  
& Blueprint returns drawings satisfying size and sheet constraints; A returns mostly policies (off-type). \\[6pt]

paligemma2-3b-mix-224 & Blueprint  
& B  
& B returns engineering drawings, including \emph{SIZE~E, 1-sheet} items; A returns documents violating allowed types. \\[6pt]

pixtral-12b & Blueprint  
& B  
& B provides engineering drawings matching the constraints; A returns documents only. \\[6pt]

Llama-4-Scout-17B-Instruct & Blueprint  
& B  
& Blueprint returns drawings with rank 2--3 satisfying \emph{SIZE~E} and \emph{1-sheet}; A returns only documents. \\

\bottomrule
\end{tabular}

\caption{Arena-style LLM-judge comparisons for the query ``drawings size E and only 1 sheet.''  
All judgments were produced by the GPT-5 arena judge. Across all baselines, \textsc{Blueprint} consistently satisfies the ENGINEERING\_DRAWING type constraint and returns valid size~E, 1-sheet drawings near the top of its ranking.
}
\label{tab:arena_examples_sizeE}
\end{table*}

\textbf{Retrieval scoring.} In a similar vein, each LLM judge was given an anonymous
model along with its top-$k=3$ retrieved documents and asked to score each document
\textit{independently} on a 3-point scale (0/1/2). All judges received identical
prompting, token limits, and formatting to ensure consistency. Importantly, retrieval
scoring was always conducted \emph{one model at a time}: each judge saw only a single
system’s top-$k$ results for independent scoring. We initially experimented with
presenting multiple systems simultaneously in a ranked list, but this introduced
strong positional bias (items shown earlier tended to receive higher scores,
regardless of model). The final protocol therefore uses single-system evaluation to
eliminate position-induced preference.

Figure~\ref{fig:llm-judge-statistical-prompt} shows the verbatim prompt given to each
judge. Table~\ref{tab:stats_examples_procedure} illustrates a representative example
of this scoring protocol. Interestingly, in this case \textsc{Blueprint}'s rank-1
result received a score of 1, whereas several baselines obtained a 2 at rank~1.
However, \textsc{Blueprint} achieved scores of 2 at both rank~2 and rank~3, while
competing models typically returned off-type or irrelevant items beyond their first
result. This highlights the advantage of evaluating all top-$k$ items independently:
although some models occasionally surface a strong item at rank~1, \textsc{Blueprint}
demonstrates stronger overall retrieval quality across the ranked list.

\begin{figure*}[t]
\centering
\begin{minipage}{0.95\textwidth}
\footnotesize
\begin{verbatim}
You are scoring engineering retrieval results for a legacy archive.

You will be given:
- the user query (it may ask for a DRAWING, a DOCUMENT, or BOTH)
- optional Allowed types (ENGINEERING_DRAWING, POLICY, PROCEDURE)
- up to 3 ranked results from ONE system

Each result includes: rank (1 is best), item_id, kind (DRAWING or DOCUMENT),
and an OCR text snippet (may include drawing numbers, revs, sheets, WBS/building,
materials, authors, dates, parts lists, etc.).

Your task:
Score EVERY result independently using the 0/1/2 scale:
0 = not relevant or clearly wrong
1 = partially relevant (some match, incomplete)
2 = fully relevant / excellent answer to the query

Rules:
- Anchors (IDs, codes) strongly support score 2.
- Slight OCR noise is acceptable.
- If a query uses negation (e.g., “no parts list”), violating items score 0.
- If unclear, be conservative (0 or 1).

Modality and constraints:
- Allowed types are a hard constraint.
- If query asks for DRAWING, score drawing cues.
- If query asks for DOCUMENT, score document cues.
- If BOTH, either modality can score 2.

Return STRICT JSON ONLY:
{
  "ratings": [
    {"item_id": "SYSTEM-RANK1", "score": 0 | 1 | 2},
    {"item_id": "SYSTEM-RANK2", "score": 0 | 1 | 2}
  ]
}
\end{verbatim}
\end{minipage}
\caption{Prompt used for per-document relevance scoring in the LLM-as-judge statistical evaluation.}
\label{fig:llm-judge-statistical-prompt}
\end{figure*}

\begin{table*}[t]
\centering
\footnotesize

\begin{tabular}{c}
\toprule
\textbf{Query:} ``retrieve the instruction manual for dissolved oxygen sensors'' \\
\bottomrule
\end{tabular}

\vspace{4pt}

\begin{tabular}{p{0.22\linewidth} p{0.18\linewidth} p{0.55\linewidth}}
\toprule
\textbf{Model} & \textbf{Scores (1/2/3)} & \textbf{Judge Explanation} \\
\midrule

Blueprint  
& 1,\;2,\;2  
& Returns procedures/manual-like documents; rank-2 and rank-3 appear strongly relevant to dissolved oxygen sensors. \\[6pt]

Llama-3.2-11B-Vision-Instruct  
& 2,\;0,\;0  
& Rank-1 matches intent (procedure), but lower ranks are irrelevant or off-type. \\[6pt]

llava-v1.6-mistral-7b-hf  
& 0,\;0,\;0  
& Returns entirely irrelevant or off-type results; no clear procedural match. \\[6pt]

paligemma2-3b-mix-224  
& 2,\;0,\;0  
& Rank-1 is a correct procedure; ranks-2/3 do not relate to the query. \\[6pt]

pixtral-12b  
& 0,\;2,\;0  
& Rank-2 is a valid procedure/manual; the others are unrelated. \\[6pt]

Llama-4-Scout-17B-16E-Instruct  
& 2,\;2,\;0  
& Strong relevant matches at ranks-1 and 2; rank-3 unrelated. \\

\bottomrule
\end{tabular}

\caption{Per-document statistical scoring (0/1/2 relevance) for the query 
``retrieve the instruction manual for dissolved oxygen sensors.''  
Each LLM judge evaluated one model at a time using the scoring prompt in 
Figure~\ref{fig:llm-judge-statistical-prompt}. Higher scores correspond to 
closer alignment with procedural intent and allowed type (PROCEDURE).}
\label{tab:stats_examples_procedure}
\end{table*}

\section{Benchmark Query Set}
Our retrieval benchmark consists of 375 anonymized queries designed to span the full
range of information needs encountered in large engineering archives. The queries
were constructed to be realistic, diverse, and balanced across modalities. Each
query is expressed in natural language, and many contain multiple constraints or
anchors (e.g., material type, revision, component count, date, hazard category,
naming pattern, or facility/project reference).

\paragraph{Query distribution.}
The 375 queries are evenly partitioned into three categories:
\begin{itemize}
    \item \textbf{150 Vision-only:} Targeting engineering drawings exclusively.
          These include geometric constraints (e.g., drawing size, sheet count),
          structural cues (e.g., electrical vs.\ mechanical vs.\ civil),
          component-based cues (e.g., part counts or manufacturer hints),
          and revision/identifier patterns.
    \item \textbf{150 Document-only:} Targeting policies, procedures, manuals, and
          other text-heavy technical documents. These queries emphasize temporal
          constraints, naming conventions, hazard references, authorship cues,
          operational context, and external standards.
    \item \textbf{50 BOTH (Multimodal):} Requiring a \emph{pair} of items, a document and a drawing, that jointly satisfy a shared anchor such as a common identifier, project code, building reference, or cross-linked operational procedure. These represent realistic workflows in engineering
    environments where drawings and procedures must be consulted together.
\end{itemize}

\paragraph{Motivation and design goals.}
The query set is intended to:
\begin{itemize}
    \item capture the heterogeneity of real-world engineering information needs,
    \item require multimodal understanding (visual + textual) for a substantial
          portion of the benchmark,
    \item stress-test both OCR quality and normalization robustness,
    \item include queries with varying specificity (broad, narrow, multi-constraint),
    \item and avoid any bias toward a particular retrieval model's strengths.
\end{itemize}

\paragraph{Anonymization.}
All identifiers, including drawing numbers, facility codes, building identifiers,
document names, and organization-specific prefixes, were replaced with abstracted
tokens or structurally similar placeholders. The semantic structure of each query
was preserved, but no proprietary or sensitive identifiers remain.

\paragraph{Ground-truth relevance.}
Each query is linked to one or more target items in the corpus, with a graded
relevance score of \texttt{0/1/2}.  
A relevance of~2 indicates a fully correct and highly informative match; a score of~1
indicates a partial or weakly correct match; and a score of~0 denotes irrelevance.
Target items were manually validated to ensure correctness and modality alignment
(e.g., vision-only queries never specify document targets).

\paragraph{Evaluation protocol.}
For each system and query, the top--3 retrieved items are judged independently
(either by human annotators or by the LLM-judge framework described in
Section~\ref{sec:retrieval}). Metrics such as nDCG@3, MAP@3, P@3, R@3, and
Success@3 are computed per query and then averaged over their respective groups
(vision, document, and multimodal).

\paragraph{Representative examples.}
Table~\ref{tab:queries_examples} shows anonymized examples capturing the range of
query intents. These illustrate the diversity of constraints (quantitative,
structural, operational, temporal) and the multimodal nature of the benchmark.

\begin{table*}[t]
\centering
\small
\setlength{\tabcolsep}{6pt}
\caption{Representative examples of the 375 retrieval queries used in the benchmark. 
All identifiers, drawing numbers, facility codes, and document names have been anonymized.
Queries cover vision-only (drawings), document-only (policies/procedures), and BOTH
(multimodal) intents.}
\label{tab:queries_examples}

\begin{tabular}{p{0.13\linewidth} p{0.82\linewidth}}
\toprule
\textbf{Type} & \textbf{Representative Query (Anonymized)} \\
\midrule

\textbf{VISION} &
``Find mechanical drawings that show parts weighing approximately 0.3 pounds.'' \\[4pt]

\textbf{VISION} &
``Retrieve engineering drawings that are size~E and contain only a single sheet.'' \\[4pt]

\textbf{VISION} &
``Retrieve a drawing revised to version~C that depicts an electrical line diagram without a parts list.'' \\[4pt]

\textbf{VISION} &
``Find structural drawings related to the concrete portion of a facility, labeled with the building’s project code.'' \\[4pt]

\textbf{VISION} &
``Retrieve drawings that contain more than 25 components and include manufacturer cues such as common industrial suppliers.'' \\
\midrule

\textbf{DOCUMENT} &
``Retrieve a procedure involving coordination with a maintenance or operability manager.'' \\[4pt]

\textbf{DOCUMENT} &
``Retrieve a procedure created in early February that follows the naming pattern ‘XYZ-\*-1.2.3’.'' \\[4pt]

\textbf{DOCUMENT} &
``Retrieve a lift-plan document that references use of an overhead bridge crane and inspection requirements.'' \\[4pt]

\textbf{DOCUMENT} &
``Retrieve a policy covering hazards such as welding, hot work, insects/wildlife, and scaffolding controls.'' \\[4pt]

\textbf{DOCUMENT} &
``Retrieve a policy prepared by an external organization and revised in the mid-2000s related to safety or compliance.'' \\
\midrule

\textbf{BOTH} &
``Retrieve BOTH the access-control procedure for a target-transfer operation AND the associated concrete building plan drawing.'' \\[4pt]

\textbf{BOTH} &
``Retrieve BOTH the facility document with identifier ending in `-AB01234` AND the drawing whose title block includes the matching facility code.'' \\[4pt]

\textbf{BOTH} &
``Retrieve BOTH the HVAC/mechanical drawing showing a drain line AND the companion ductwork document referenced by that design.'' \\[4pt]

\textbf{BOTH} &
``Retrieve BOTH a policy with a specific project code AND the drawing whose title block displays the same project identifier.'' \\[4pt]

\textbf{BOTH} &
``Retrieve BOTH a feeder-schedule document AND the site-utilities plan drawing referenced within it.'' \\

\bottomrule
\end{tabular}
\end{table*}

\end{document}